\documentclass[11pt,a4paper]{article}

\usepackage[T1]{fontenc}
\usepackage[utf8]{inputenc}
\usepackage{lmodern}
\usepackage[margin=2.5cm]{geometry}
\usepackage{amsmath,amssymb}
\usepackage{graphicx}
\usepackage{booktabs}
\usepackage{tabularx}
\usepackage{array}
\usepackage{multirow}
\usepackage{subcaption}
\usepackage[authoryear,round]{natbib}
\usepackage[hidelinks]{hyperref}
\usepackage{xcolor}
\usepackage{authblk}

\newcolumntype{Y}{>{\raggedright\arraybackslash}X}

\title{Benchmarking Classical Machine Learning Models for Shading-Fault Detection in Rooftop Photovoltaic Systems Using Real Sensor Data}

\author[1]{Zablon Ombiro\thanks{Corresponding author: Zablon Ombiro, \href{mailto:zablon.ombiro@strathmore.edu}{zablon.ombiro@strathmore.edu}}}
\author[1]{Bernard Shibwabo}
\affil[1]{School of Computing and Engineering Sciences, Strathmore University, Nairobi, Kenya}
\date{}

\begin{document}
\maketitle

\begin{abstract}
Most published comparisons of machine-learning classifiers for photovoltaic (PV) fault detection are validated on simulated faults. We compare seven classical and ensemble-tree classifiers on the R\'eunion dataset, a 4\,kW domestic rooftop plant on which shading faults were physically staged and logged. Readings within a staged episode are near-duplicates, so every train/test split is grouped by episode. A 20{,}800-row subsample that retains all 17 fault episodes was evaluated with three repeats of five-fold group-aware nested cross-validation, with synthetic minority oversampling (SMOTE) applied inside the training folds. Random Forest and XGBoost gave the highest mean accuracy (0.838 and 0.831) and Matthews correlation coefficient (MCC; 0.526 and 0.522). After Holm correction across the 21 pairwise Wilcoxon tests, only Random Forest's accuracy advantage over k-nearest neighbours and the multi-layer perceptron remained significant. Logistic Regression and the Decision Tree (accuracy 0.820) could be separated neither from the ensembles nor from the weakest models. SMOTE lowered binary MCC for six of the seven classifiers. Errors were concentrated in whole episodes: the Random Forest almost entirely missed four of the 17 fault episodes, including one of two uniform-shading and two of five intermittent-shading episodes. A post-hoc diagnostic shows that the no-fault rows were sampled over a wider range of times of day and output levels than the fault rows. Clock time alone reaches an MCC of 0.44, and PV-side voltage alone 0.72. The reported scores therefore reflect operating-condition differences as well as shading signatures. We release the pipeline and the diagnostics, and identify matched no-fault sampling as the step needed before the classifier ranking can be generalised.
\end{abstract}

\noindent\textbf{Keywords:} photovoltaic fault detection; shading faults; classifier comparison; group-aware cross-validation; class imbalance; SHAP

\section{Introduction}

Partial or uniform shading forces shaded cells into reverse bias. They then dissipate rather than generate power, and the resulting hot spots are a recognised fire-risk pathway. On small rooftop systems the loss is often invisible on a production dashboard, so detection has to come from the inverter's electrical measurements.

Classical and ensemble-tree classifiers are the usual baselines for this kind of tabular monitoring data, and several multi-model comparisons exist \citep{Eldeghady2024,Khandeparkar2025,Mouleloued2025}. Most of them use simulated faults, because fault-labelled field data are scarce. Field data bring a problem that simulation hides. A staged fault persists for minutes to hours, so adjacent readings are almost identical, and a random row-level split puts near-copies of the same reading on both sides of the train/test boundary. Rankings obtained under such splits, or on simulated data, cannot be assumed to hold under field conditions. The closest real-data comparison we know of \citep{TocheTchio2024} does not group its splits and does not test differences between classifiers.

We use the R\'eunion dataset \citep{Graillet2022}, one of the few public records in which shading faults were staged and logged on an operating plant. Seven classifiers are compared under a protocol that keeps every episode wholly inside one fold, and the pairwise differences are tested with multiplicity correction. Four questions guide the analysis:
\begin{itemize}
\item \textbf{RQ1.} Under episode-grouped evaluation, which classifiers perform best, and which differences survive statistical testing?
\item \textbf{RQ2.} Does SMOTE, applied inside the training folds, change the ranking or only the absolute scores, and does its effect depend on the model family?
\item \textbf{RQ3.} Which electrical signals drive the decisions, and is that consistent across model families?
\item \textbf{RQ4.} Do the less interpretable models give a measurable gain over Logistic Regression and a single Decision Tree?
\end{itemize}

The contributions are: (i) a seven-classifier benchmark on real staged shading faults; (ii) an episode-grouped nested evaluation, made tractable by documented subsampling; (iii) multi-metric results with Holm-corrected paired tests and effect sizes, and a per-model normalised SHAP comparison; and (iv) episode-level error analysis and confound diagnostics that bound how far the results generalise. The pipeline, results and diagnostics are public.

\section{Related work}

\citet{Eldeghady2024}, \citet{Khandeparkar2025} and \citet{Mouleloued2025} compare supervised classifiers, including Random Forest and, in one case, XGBoost, on simulated PV fault data. They report point estimates without cross-model significance tests. Ensembles have also been validated on real DC-side and multi-plant data \citep{Chen2018,Dhibi2021,Harrou2022,Hussain2022}, including about 70{,}000 readings from a 500\,kWp system \citep{Gaaloul2025}. Related comparisons pit the support vector machine (SVM) against XGBoost \citep{Suliman2024} and combine logistic regression, decision-tree and SVM learners in a voting ensemble \citep{Yang2022}. \citet{Mellit2022} find that the advantage of ensembles over classical learners is real but not universal, which is the case for testing differences rather than reading them off point estimates. \citet{TocheTchio2024} compare Extra Trees with six classifiers (Logistic Regression, kNN, SVM, decision tree, AdaBoost and Random Forest) on low-cost-sensor PV data that include partial shading, without grouped splits or significance testing. Class imbalance and label noise have been handled with RUSBoost \citep{Adhya2023} and distributionally robust logistic regression \citep{Wang2023}. Ensembles have also been applied to thermal images \citep{KrishnaVeni2025} and to an emulator testbed \citep{QuilesCucarella2025}.

Real field data have mostly been used for other tasks. \citet{Kappler2024} combine a long short-term memory network with a one-diode model for shading-aware power forecasting. PVF-10 provides real drone thermal imagery \citep{Wang2024a}. \citet{Graillet2022} describe the dataset used here but report no classifier.

The classifiers span distinct inductive biases: k-nearest neighbours (kNN; \citealp{Cover1967}), the maximum-margin SVM \citep{Cortes1995}, the decision tree \citep{Quinlan1986}, the multi-layer perceptron (MLP; \citealp{Rumelhart1986}), logistic regression as the linear reference, and the bagged and boosted ensembles Random Forest \citep{Breiman2001} and XGBoost \citep{Chen2016}. A deep sequence model trained on simulated data \citep{Amiri2024} motivates the sequence-aware extension left to future work.

Native importance measures, such as a logistic coefficient or a tree's impurity reduction, cannot be compared across model families. SHAP \citep{Lundberg2017} is the usual remedy in PV fault detection \citep{Utama2023} and in related energy and machinery studies \citep{Cakiroglu2024,Santos2024,Bindingso2023}. It is comparable across models only when the explained outputs share a scale, a point we return to in Section~\ref{sec:shap}.

Accuracy is misleading under class imbalance. The Matthews correlation coefficient (MCC) is symmetric in classes and error types \citep{Chicco2020}. SMOTE \citep{Chawla2002} is the reference oversampler, but its benefit depends on the classifier \citep{Swana2022,MaldonadoCorrea2024}, and in transformer diagnosis it has mainly helped tree ensembles \citep{Prasojo2023,Wang2024b}. For comparing classifiers we follow \citet{Demsar2006}. All-pairwise testing requires a multiplicity correction \citep{Garcia2008}. Effect sizes \citep{Kerby2014} complement $p$-values, although they do not settle the wider critique of null-hypothesis testing \citep{Benavoli2017}. \citet{Wainer2021} show that non-nested tuning is usually adequate in practice, so the nested design used here is the conservative choice.

Table~\ref{tab:related} positions this study against the closest prior work.

\begin{table}[t]
\centering
\caption{Closest prior studies and how this work differs.}
\label{tab:related}
\small
\begin{tabularx}{\textwidth}{@{}p{2.9cm}YY@{}}
\toprule
Study & Data and task & Difference in this work \\
\midrule
\citet{Eldeghady2024} & Simulated PV array faults; several supervised classifiers; accuracy-focused & Real staged faults; six metrics; corrected paired tests \\
\citet{Khandeparkar2025} & Simulated fault signatures, four types; five models incl.\ RF and XGBoost & Same ensemble baselines on real data with grouped splits \\
\citet{Mouleloued2025} & Seven-class fault dataset; five models incl.\ RF & MCC and AUC-ROC alongside accuracy; significance testing \\
\citet{TocheTchio2024} & Low-cost-sensor data incl.\ partial shading; Extra Trees vs.\ six classifiers & Episode-grouped CV, repeated paired tests, SHAP \\
\citet{Utama2023} & Comparison of explanation methods for PV faults & SHAP across seven classifiers on a normalised scale \\
\citet{Kappler2024} & Real installation; LSTM + physical model; forecasting & Discrete fault classification; multi-model comparison \\
\citet{Graillet2022} & Data descriptor of the R\'eunion plant & First classifier benchmark on this record \\
\bottomrule
\end{tabularx}
\end{table}

\section{Methodology}

\subsection{Dataset and target construction}

We use the electrical-production log of inverter~1 of the R\'eunion plant \citep{Graillet2022}: a 4\,kW, 24-module installation at 26$^\circ$ tilt in a tropical climate. The log has one sample every 5--8\,s and seven channels: PV-side current $I_a$, grid current $I_g$, daily cumulative grid energy $E_g$, grid frequency $F_g$, grid power $P_g$, PV-side voltage $V_a$ and grid voltage $V_g$. Faults were staged and logged with an explicit code, so the labels record an experimental condition rather than an inference.

After rows with missing values are dropped, the file has 2{,}988{,}798 rows from 1 October 2021 to 13 April 2023 (559 days). Table~\ref{tab:codes} gives the fault-code distribution. No-fault rows make up 97.17\% of the record. Splitting the series at every change of fault state gives 35 contiguous runs: 17 fault episodes and 18 no-fault runs. All fault episodes fall between 8 February and 19 May 2022, a window of about 100 days.

\begin{table}[t]
\centering
\caption{Fault-code distribution of the full record (2{,}988{,}798 rows), before subsampling.}
\label{tab:codes}
\small
\begin{tabular}{@{}llrrc@{}}
\toprule
Code & Description & Rows & Share (\%) & Episodes \\
\midrule
0.0 & No fault & 2{,}904{,}284 & 97.17 & 18 \\
1.0 & Uniform shading (entire string obscured) & 6{,}318 & 0.21 & 2 \\
2.1 & Constant partial shading, 1 module & 2{,}148 & 0.07 & 1 \\
2.2 & Constant partial shading, 2 modules & 2{,}460 & 0.08 & 1 \\
2.3 & Constant partial shading, 3 modules & 3{,}287 & 0.11 & 1 \\
3.1 & Constant partial shading, 1/3 module & 2{,}717 & 0.09 & 3 \\
3.2 & Constant partial shading, 2/3 module & 6{,}292 & 0.21 & 4 \\
4.0 & Intermittent partial shading (moving shadow) & 61{,}292 & 2.05 & 5 \\
\bottomrule
\end{tabular}
\end{table}

Nested tuning of seven classifiers on about three million rows is not feasible on a single CPU workstation. The pipeline therefore keeps a random contiguous slice of at most 400 rows per fault episode and 2{,}000 rows per no-fault run, and drops whole no-fault runs at random until no more than 15{,}000 no-fault rows remain. All 17 fault episodes are retained. The benchmark has 20{,}800 rows in 24 episodes: 14{,}000 no-fault, 4{,}000 constant-partial-shading and 2{,}800 severe-tier rows, a fault prevalence of 32.7\%. A constant no-fault predictor scores 0.673 accuracy (MCC~0) on it. All metrics below describe this capped benchmark, not the 2.83\%-prevalence record.

The binary label is read directly from the fault code ($y=0$ for code 0.0, otherwise $y=1$). A three-tier label groups codes by extent, following \citet[Table~3]{Graillet2022}: tier~0, no fault; tier~1, constant partial shading (2.1--3.2); and tier~2, severe shading (1.0 uniform and 4.0 intermittent). The grouping is a modelling choice, which is why both schemes are reported.

\subsection{Features and class-imbalance handling}

All 29 features are computed within an episode, so no window crosses an episode boundary. There are four groups:
\begin{itemize}
\item the seven raw channels;
\item four instantaneous features: $I_g/I_a$, $P_g/(V_aI_a)$, $|F_g-50|$ and $V_g-V_a$;
\item sixteen rolling features: the mean $\mu_w$ and standard deviation $\sigma_w$ of $P_g$, $I_a$, $V_g$ and $F_g$ over $w=5$ and $10$ samples (about 30 and 60\,s);
\item two first differences, $\Delta P_g$ and $\Delta I_a$.
\end{itemize}
Because $V_g$ stays near 233\,V whenever the inverter is connected (s.d.\ about 4\,V over no-fault rows), $V_g-V_a$ carries essentially the same information as $V_a$.

Logistic Regression, kNN, the SVM and the MLP receive features standardised on the training fold; the tree models use raw scales. SMOTE (imbalanced-learn, three neighbours) is fitted inside each training fold, including the inner tuning folds. Every run is repeated without SMOTE.

\subsection{Classifiers and tuning}

Table~\ref{tab:space} lists the search spaces. Hyperparameters are chosen by randomised search (15 draws, or every configuration where the space is smaller, as for Logistic Regression's six) inside each outer training fold. The search uses an inner three-fold group-aware split scored on macro-F1, so the outer test fold never informs selection. Table~\ref{tab:modal} gives the configuration selected most often under the primary condition. The modal choices are strongly regularised: the smallest C for Logistic Regression, depth~3 for the Decision Tree and XGBoost, and a minimum leaf size of 10 for Random Forest. This is consistent with only 17 distinct fault episodes being available for learning.

\begin{table}[t]
\centering
\caption{Randomised-search space per model ($d$ = number of features; $q$ = Minkowski exponent).}
\label{tab:space}
\small
\begin{tabularx}{\textwidth}{@{}lY@{}}
\toprule
Model & Search space \\
\midrule
Logistic Regression & $C\in\{10^{-3},10^{-2},10^{-1},1,10,100\}$; L2 penalty (all 6 evaluated) \\
kNN & $k\in\{3,5,11,21,31,51\}$; weights $\in$ \{uniform, distance\}; $q\in\{1,2\}$ \\
Decision Tree & max depth $\in\{3,4,6,8,12,16,\infty\}$; min samples per leaf $\in\{1,5,10,20\}$; criterion $\in$ \{Gini, entropy\} \\
SVM (RBF) & $C\in\{0.1,1,10,100\}$; $\gamma\in$ \{scale, auto, $10^{-3}$, $10^{-2}$, $10^{-1}$\} \\
MLP & hidden layers $\in$ \{(16), (32), (64), (16,8), (32,16)\}; $\alpha\in\{10^{-4},\dots,10^{-1}\}$; initial learning rate $\in\{5\times10^{-4},10^{-3},10^{-2}\}$; early stopping \\
Random Forest & trees $\in\{100,200,300\}$; max depth $\in\{\infty,8,16\}$; min samples per leaf $\in\{1,5,10\}$; features per split $\in\{\sqrt d,\log_2 d\}$ \\
XGBoost & trees $\in\{100,300,500\}$; max depth $\in\{3,6,9\}$; learning rate $\in\{0.01,0.05,0.1,0.3\}$; row and column subsampling $\in\{0.7,0.85,1.0\}$ \\
\bottomrule
\end{tabularx}
\end{table}

\begin{table}[t]
\centering
\caption{Modal configuration across the 15 outer-fold searches (binary scheme, with SMOTE).}
\label{tab:modal}
\small
\begin{tabularx}{\textwidth}{@{}lY@{}}
\toprule
Model & Modal configuration \\
\midrule
Logistic Regression & $C=0.001$; L2 \\
kNN & $k=3$; $q=1$; uniform weights \\
Decision Tree & max depth 3; min samples per leaf 1; Gini \\
SVM (RBF) & $C=0.1$; $\gamma=0.01$ \\
MLP & one hidden layer of 64 units; $\alpha=10^{-4}$; initial learning rate $5\times10^{-4}$ \\
Random Forest & 300 trees; max depth 8; min samples per leaf 10; $\log_2 d$ features per split \\
XGBoost & 100 trees; max depth 3; learning rate 0.01; row subsampling 0.7; column subsampling 0.85 \\
\bottomrule
\end{tabularx}
\end{table}

\subsection{Evaluation and statistical testing}

The outer loop is five-fold cross-validation, stratified by class and grouped by episode, and repeated three times with different shuffles. This gives 15 outer folds. Five folds is the largest number that leaves several fault episodes on each side of every split. Within a repeat, all 28 runs (7 classifiers $\times$ 2 label schemes $\times$ 2 resampling conditions) share the same outer partitions.

The binary scheme with SMOTE is the primary condition. We report accuracy, fault-class precision, recall and F1 (macro-averaged for the tier scheme), MCC and AUC-ROC. MCC is the headline metric, because accuracy is inflated by the 67.3\% majority class:
\begin{equation}
\mathrm{MCC}=\frac{TP\cdot TN-FP\cdot FN}{\sqrt{(TP+FP)(TP+FN)(TN+FP)(TN+FN)}}.
\end{equation}

Pairwise differences in accuracy and MCC are tested with the Wilcoxon signed-rank test on the 15 paired folds. Holm's step-down correction is applied across the 21 pairs of each metric. The matched-pairs rank-biserial correlation $r$ \citep{Kerby2014} is reported as the effect size. The three repeats reuse the same 17 fault episodes, so the 15 folds are not independent and the uncorrected tests are anti-conservative. The Holm-adjusted results should therefore be read as the more reliable ones.

\subsection{Implementation}

All experiments ran on a single CPU-only workstation, with a fixed seed (42) for SMOTE, the classifiers and the search. The outer split of repeat $i$ uses seed $42+i$. The 28 runs took 26{,}666\,s (7.4\,h), and the separate SHAP computation about 1.5--2\,h, dominated by kNN, the SVM and the MLP, for which no closed-form SHAP algorithm exists. Table~\ref{tab:env} lists the software and hardware. The post-hoc diagnostics in Sections~\ref{sec:perfault} and~\ref{sec:confound} use the same outer splits and the fixed modal Random Forest configuration.

\begin{table}[t]
\centering
\caption{Software and hardware environment.}
\label{tab:env}
\small
\begin{tabularx}{\textwidth}{@{}lY@{}}
\toprule
Component & Version \\
\midrule
Python & 3 (author's workstation) \\
pandas & $\geq$ 2.0 \\
NumPy & $\geq$ 1.24 \\
scikit-learn & $\geq$ 1.3 \\
imbalanced-learn & $\geq$ 0.11 \\
SciPy (Wilcoxon testing) & $\geq$ 1.10 \\
matplotlib & $\geq$ 3.7 \\
XGBoost & $\geq$ 2.0 (ensemble classifier, Section 3.3) \\
SHAP & $\geq$ 0.44 (feature attribution, Section 4.4) \\
Hardware & CPU only; all classifiers run unmodified on commodity laptop hardware; no GPU required \\
\bottomrule
\end{tabularx}
\end{table}

\section{Results}

\subsection{Primary comparison}

\begin{figure}[t]
\centering
\includegraphics[width=\textwidth]{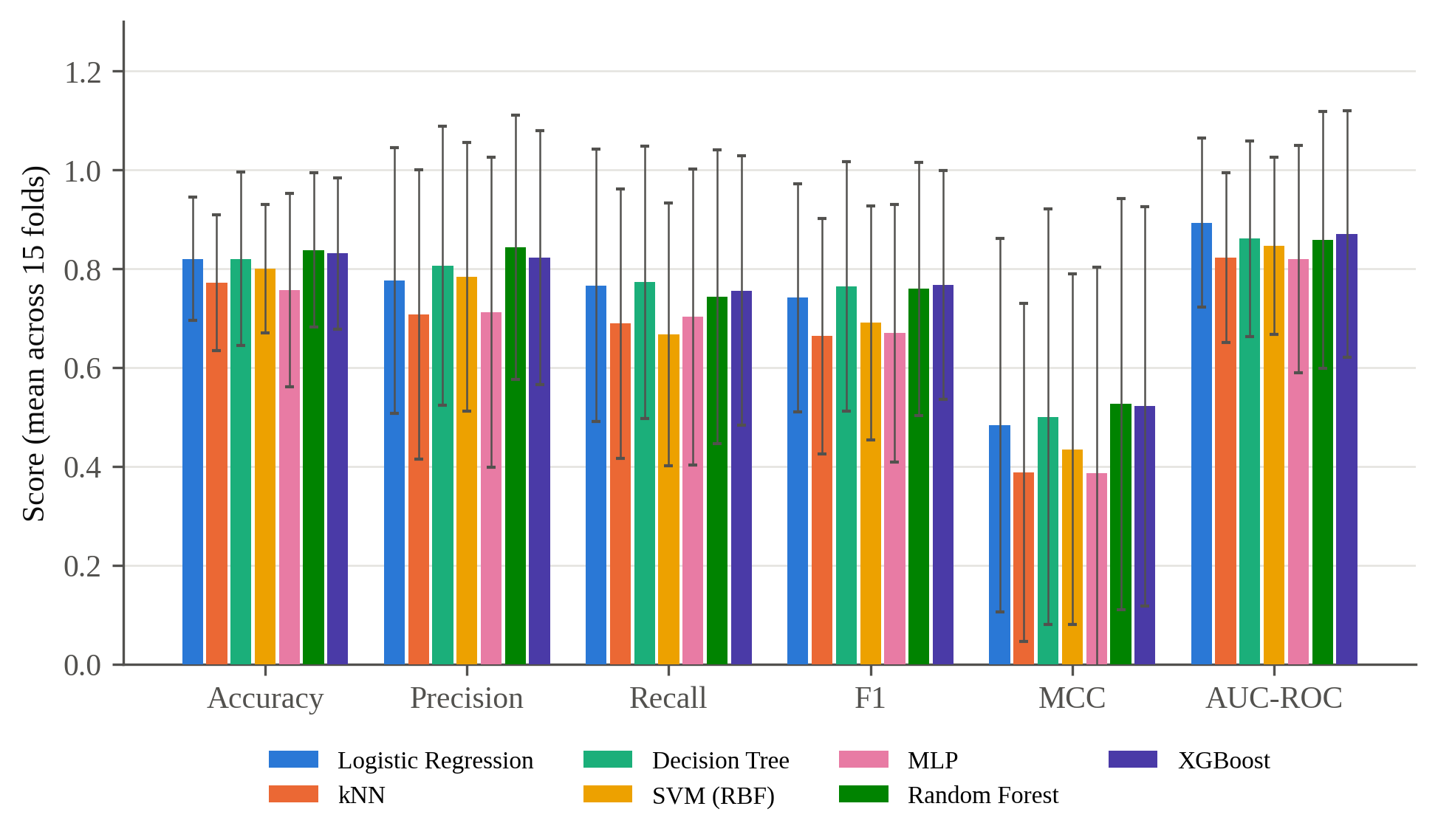}
\caption{Six metrics per classifier (binary scheme, with SMOTE); error bars show the fold-wise standard deviation.}
\label{fig:metrics}
\end{figure}

Table~\ref{tab:primary} and Fig.~\ref{fig:metrics} report the primary condition. Random Forest has the highest accuracy (0.838) and MCC (0.526), followed by XGBoost (0.831, 0.522). Logistic Regression and the Decision Tree both reach 0.820 accuracy (MCC 0.483 and 0.500). kNN, the SVM and the MLP trail (0.772, 0.800 and 0.757). Logistic Regression has the highest AUC-ROC (0.893), so its ranking of samples is good even though its default-threshold accuracy is lower. Fold-to-fold standard deviations are large for every model (Random Forest MCC 0.526 $\pm$ 0.416), because folds differ in which episodes they hold out (Section~\ref{sec:perfault}).

\begin{table}[t]
\centering
\caption{Binary scheme with SMOTE; mean $\pm$ s.d.\ over 15 outer folds.}
\label{tab:primary}
\small
\begin{tabular}{@{}lcccccc@{}}
\toprule
Model & Accuracy & Precision & Recall & F1 & MCC & AUC-ROC \\
\midrule
Logistic Regression & .820$\pm$.124 & .776$\pm$.269 & .766$\pm$.275 & .741$\pm$.231 & .483$\pm$.378 & .893$\pm$.170 \\
kNN & .772$\pm$.137 & .707$\pm$.293 & .689$\pm$.273 & .663$\pm$.238 & .388$\pm$.342 & .822$\pm$.172 \\
Decision Tree & .820$\pm$.175 & .806$\pm$.282 & .773$\pm$.276 & .764$\pm$.252 & .500$\pm$.420 & .861$\pm$.198 \\
SVM (RBF) & .800$\pm$.130 & .784$\pm$.272 & .667$\pm$.266 & .690$\pm$.237 & .435$\pm$.354 & .846$\pm$.179 \\
MLP & .757$\pm$.196 & .712$\pm$.314 & .702$\pm$.299 & .670$\pm$.261 & .387$\pm$.416 & .820$\pm$.230 \\
Random Forest & .838$\pm$.156 & .844$\pm$.267 & .744$\pm$.297 & .759$\pm$.256 & .526$\pm$.416 & .859$\pm$.260 \\
XGBoost & .831$\pm$.153 & .822$\pm$.257 & .755$\pm$.272 & .767$\pm$.232 & .522$\pm$.404 & .870$\pm$.250 \\
\bottomrule
\end{tabular}
\end{table}

Without SMOTE (Table~\ref{tab:nosmote}), Random Forest again leads (0.839, MCC 0.531), and Logistic Regression (0.830) moves ahead of XGBoost (0.814). Removing SMOTE raises accuracy for every model except XGBoost.

\begin{table}[t]
\centering
\caption{Binary scheme without SMOTE; mean $\pm$ s.d.\ over 15 outer folds.}
\label{tab:nosmote}
\small
\begin{tabular}{@{}lcccccc@{}}
\toprule
Model & Accuracy & Precision & Recall & F1 & MCC & AUC-ROC \\
\midrule
Logistic Regression & .830$\pm$.149 & .824$\pm$.265 & .742$\pm$.289 & .751$\pm$.248 & .511$\pm$.406 & .901$\pm$.170 \\
kNN & .792$\pm$.146 & .763$\pm$.278 & .693$\pm$.288 & .687$\pm$.239 & .439$\pm$.362 & .849$\pm$.177 \\
Decision Tree & .823$\pm$.189 & .818$\pm$.290 & .778$\pm$.278 & .773$\pm$.261 & .516$\pm$.442 & .878$\pm$.206 \\
SVM (RBF) & .805$\pm$.141 & .819$\pm$.268 & .657$\pm$.261 & .700$\pm$.234 & .456$\pm$.363 & .841$\pm$.175 \\
MLP & .788$\pm$.180 & .763$\pm$.294 & .722$\pm$.291 & .709$\pm$.252 & .451$\pm$.404 & .854$\pm$.220 \\
Random Forest & .839$\pm$.157 & .852$\pm$.263 & .739$\pm$.296 & .761$\pm$.255 & .531$\pm$.416 & .861$\pm$.253 \\
XGBoost & .814$\pm$.171 & .824$\pm$.287 & .727$\pm$.269 & .747$\pm$.249 & .508$\pm$.415 & .846$\pm$.258 \\
\bottomrule
\end{tabular}
\end{table}

\subsection{Confusion matrices}

\begin{figure}[t]
\centering
\begin{subfigure}{\textwidth}\centering
\includegraphics[width=0.95\textwidth]{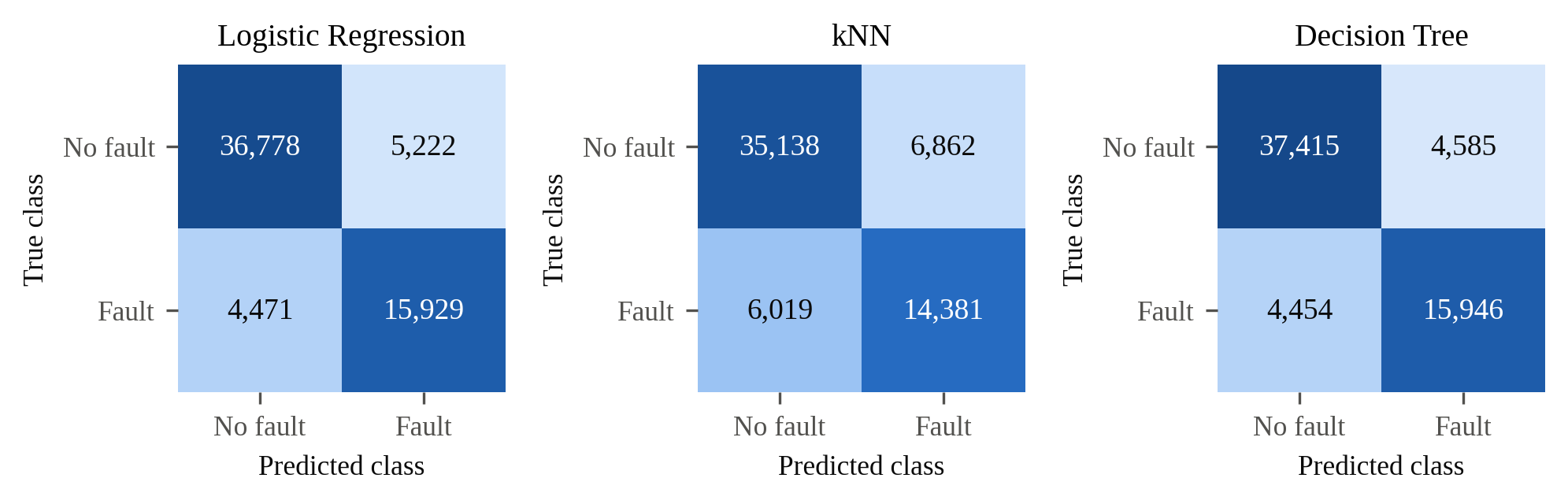}\caption{}\end{subfigure}
\begin{subfigure}{0.49\textwidth}\centering
\includegraphics[width=\textwidth]{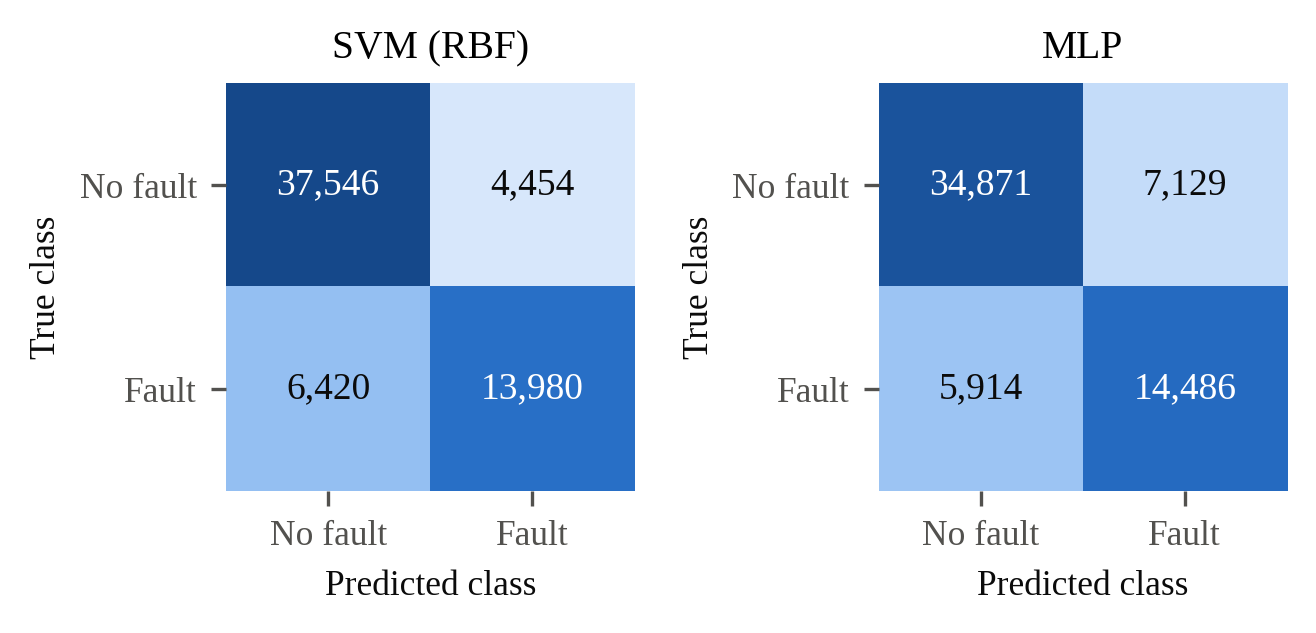}\caption{}\end{subfigure}
\begin{subfigure}{0.49\textwidth}\centering
\includegraphics[width=\textwidth]{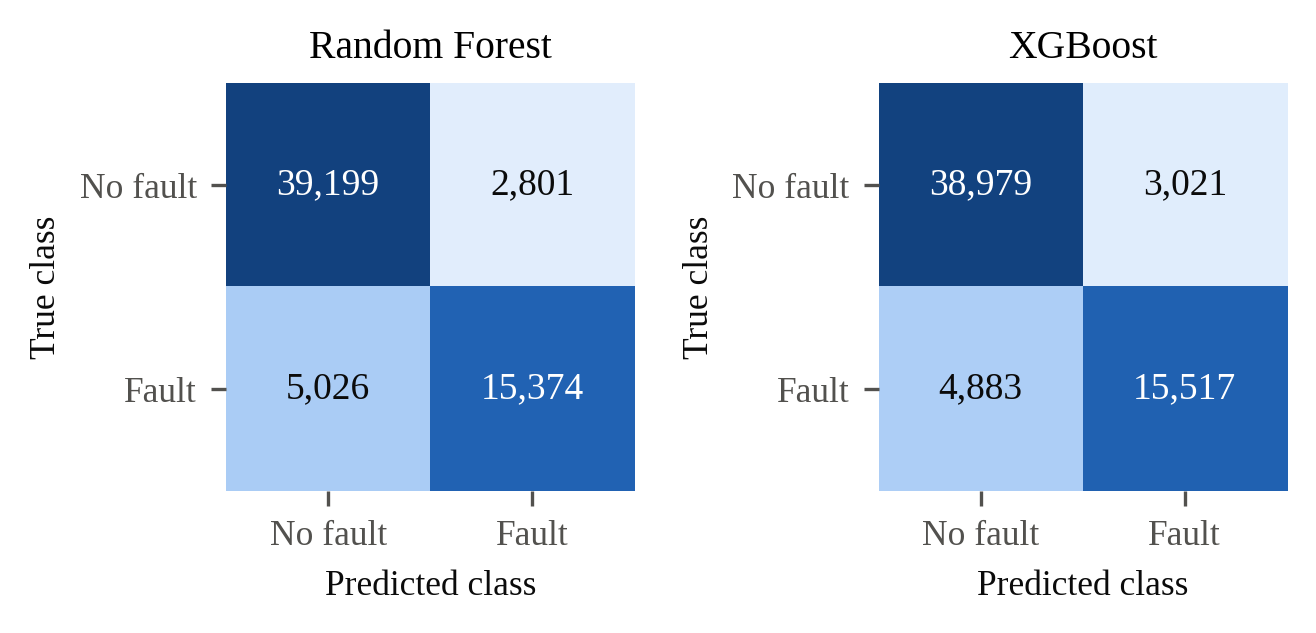}\caption{}\end{subfigure}
\caption{Pooled confusion matrices (binary, with SMOTE): (a) Logistic Regression, kNN and Decision Tree; (b) SVM and MLP; (c) Random Forest and XGBoost.}
\label{fig:cm}
\end{figure}

Table~\ref{tab:cm} and Fig.~\ref{fig:cm} pool the out-of-fold predictions: 62{,}400 per model (20{,}800 rows $\times$ 3 repeats). Random Forest and XGBoost produce the fewest false alarms (2{,}801 and 3{,}021). The SVM misses the most faults (6{,}420 false negatives; recall 0.667). Missing a fault is the costlier error when the hazard is undetected shading.

\begin{table}[t]
\centering
\caption{Pooled confusion-matrix counts (binary, with SMOTE; positive class = fault).}
\label{tab:cm}
\small
\begin{tabular}{@{}lrrrr@{}}
\toprule
Model & TN & FP & FN & TP \\
\midrule
Logistic Regression & 36{,}778 & 5{,}222 & 4{,}471 & 15{,}929 \\
kNN & 35{,}138 & 6{,}862 & 6{,}019 & 14{,}381 \\
Decision Tree & 37{,}415 & 4{,}585 & 4{,}454 & 15{,}946 \\
SVM (RBF) & 37{,}546 & 4{,}454 & 6{,}420 & 13{,}980 \\
MLP & 34{,}871 & 7{,}129 & 5{,}914 & 14{,}486 \\
Random Forest & 39{,}199 & 2{,}801 & 5{,}026 & 15{,}374 \\
XGBoost & 38{,}979 & 3{,}021 & 4{,}883 & 15{,}517 \\
\bottomrule
\end{tabular}
\end{table}

\subsection{Statistical significance}

Table~\ref{tab:wilcoxon} gives the uncorrected Wilcoxon $p$-values on accuracy. Before correction, Random Forest exceeds kNN ($p<.001$, $r=.93$) and the MLP ($p<.001$, $r=1.0$), and XGBoost exceeds both ($p=.009$, $r=.79$; $p=.010$, $r=.73$). On MCC, both ensembles exceed kNN ($p=.003$, $r=1.0$), the MLP ($p=.006$ and $.026$) and the SVM ($p=.021$ and $.016$), and the Decision Tree exceeds kNN ($p=.041$). No comparison involving Logistic Regression is significant on either metric. Neither is any comparison among Logistic Regression, the Decision Tree and the two ensembles (all $p\geq.29$ on MCC and $p\geq.42$ on accuracy).

After Holm correction over the 21 pairs, two comparisons survive on accuracy: Random Forest over the MLP ($p_{\mathrm{Holm}}=.001$) and over kNN ($p_{\mathrm{Holm}}=.009$). None survives on MCC; the smallest is $p_{\mathrm{Holm}}=.070$, for both ensembles versus kNN. The large effect sizes show that the ensembles beat kNN and the MLP in most folds. But with 17 fault episodes and correlated repeats, the design can confirm only the largest gaps. The lack of significant differences among Logistic Regression, the Decision Tree and the ensembles is a failure to detect a difference. It does not show that they perform equally.

\begin{table}[t]
\centering
\caption{Pairwise Wilcoxon signed-rank $p$-values on accuracy (15 paired folds; upper triangle). Bold: significant after Holm correction over 21 pairs. $r$: matched-pairs rank-biserial correlation, reported where $p<.05$ uncorrected; positive values favour the column model.}
\label{tab:wilcoxon}
\small
\begin{tabular}{@{}lcccccc@{}}
\toprule
 & kNN & DT & SVM & MLP & RF & XGB \\
\midrule
LR & .055 & .599 & .055 & .277 & .720 & .847 \\
kNN & & .151 & .188 & 1.000 & \textbf{$<$.001} ($r$=.93) & .009 ($r$=.79) \\
DT & & & .303 & .095 & .679 & .975 \\
SVM & & & & .421 & .121 & .121 \\
MLP & & & & & \textbf{$<$.001} ($r$=1.0) & .010 ($r$=.73) \\
RF & & & & & & .421 \\
\bottomrule
\end{tabular}
\end{table}

\subsection{Feature attribution}
\label{sec:shap}

\begin{figure}[t]
\centering
\includegraphics[width=0.82\textwidth]{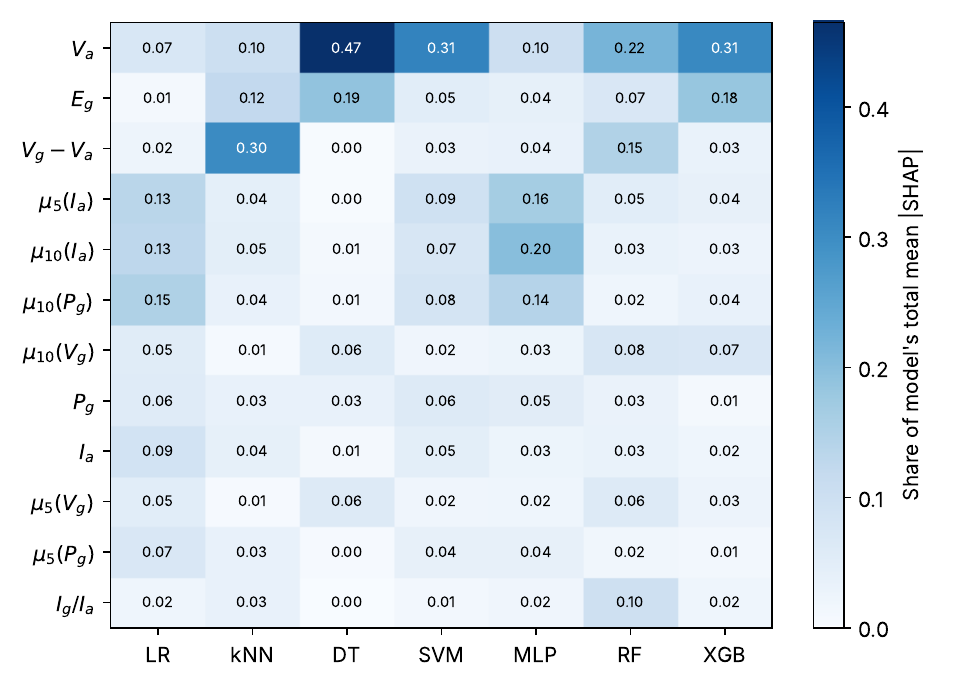}
\caption{Share of each model's total mean $|$SHAP$|$ for the twelve features with the largest average share (binary, with SMOTE). $\mu_w$: $w$-sample rolling mean.}
\label{fig:shap}
\end{figure}

SHAP values were computed for every classifier fitted on the full benchmark (binary, with SMOTE). Two caveats apply. First, the explained models use library-default hyperparameters, not the tuned configurations of Table~\ref{tab:modal}, and the attributions are in-sample. Second, the explainers return different output scales: log-odds for Logistic Regression and XGBoost, probability for the others. Raw mean $|$SHAP$|$ values therefore cannot be compared or summed across models. Fig.~\ref{fig:shap} normalises each model's attributions to shares of its own total.

$V_a$ ranks first for the Decision Tree, Random Forest, XGBoost and the SVM, third for kNN and fourth for the MLP. kNN's top feature, $V_g-V_a$, is itself a near-linear transform of $V_a$. Logistic Regression relies mostly on rolling means of $P_g$ and $I_a$ and ranks $V_a$ fifth. Averaged over the seven models, $V_a$ accounts for 23\% of the attribution, followed by $E_g$ (9\%) and $V_g-V_a$ (8\%). $E_g$ is a daily cumulative counter, so its prominence is a warning sign rather than a physical finding (Section~\ref{sec:confound}).

\subsection{Label scheme and resampling}

\begin{figure}[t]
\centering
\begin{subfigure}{0.49\textwidth}\centering
\includegraphics[width=\textwidth]{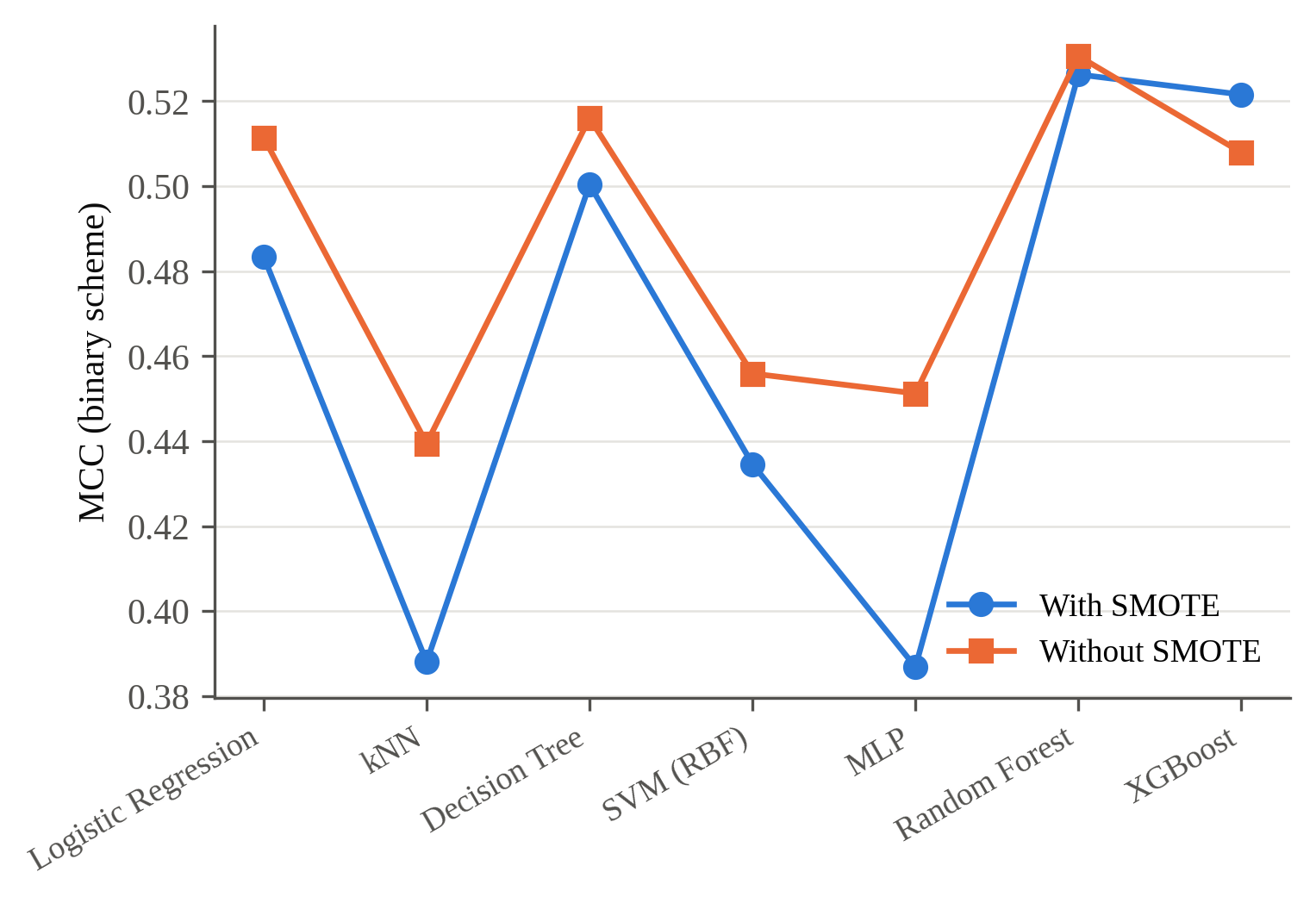}\caption{}\label{fig:smote}\end{subfigure}
\begin{subfigure}{0.49\textwidth}\centering
\includegraphics[width=\textwidth]{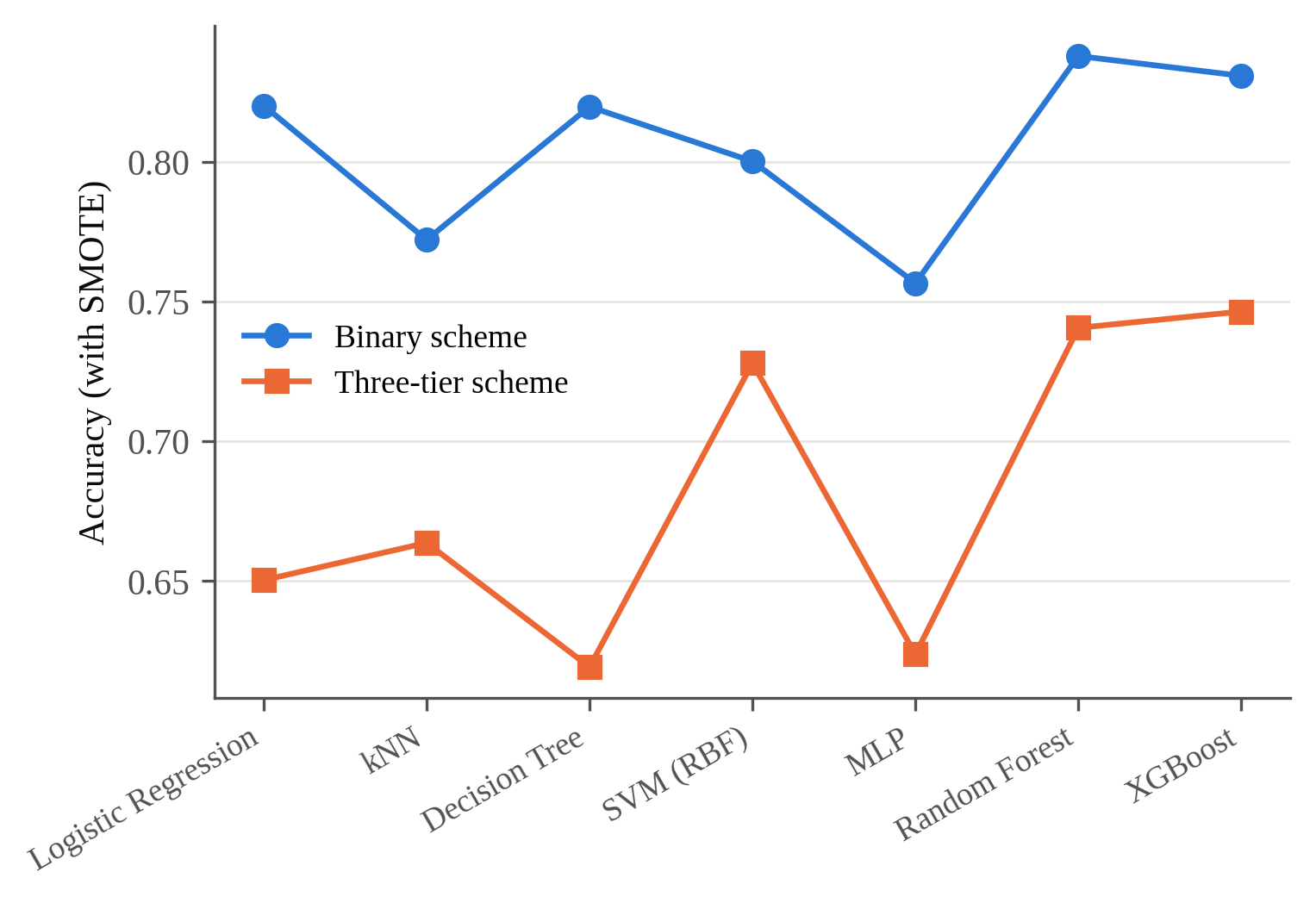}\caption{}\label{fig:tier}\end{subfigure}
\caption{(a) Effect of SMOTE on binary MCC. (b) Accuracy under the binary and three-tier schemes, with SMOTE.}
\end{figure}

Table~\ref{tab:tier} compares the binary and three-tier schemes under both resampling conditions. With SMOTE, every classifier loses accuracy and MCC on the tier task, and the ensembles keep the highest tier scores (XGBoost 0.746 / 0.477; Random Forest 0.741 / 0.465). Without SMOTE, the order changes: the Decision Tree (0.763 / 0.527) and Logistic Regression (0.758 / 0.483) lead, and the Decision Tree loses only 6 accuracy points from binary to tier, against 20 with SMOTE. The SVM's apparent robustness to the finer task (Fig.~\ref{fig:tier}) is therefore specific to the SMOTE condition.

\begin{table}[t]
\centering
\caption{Binary versus three-tier scheme, with and without SMOTE (mean over 15 folds).}
\label{tab:tier}
\small
\begin{tabular}{@{}lcccccccc@{}}
\toprule
 & \multicolumn{4}{c}{With SMOTE} & \multicolumn{4}{c}{Without SMOTE} \\
\cmidrule(lr){2-5}\cmidrule(l){6-9}
 & \multicolumn{2}{c}{Accuracy} & \multicolumn{2}{c}{MCC} & \multicolumn{2}{c}{Accuracy} & \multicolumn{2}{c}{MCC} \\
Model & bin. & tier & bin. & tier & bin. & tier & bin. & tier \\
\midrule
Logistic Regression & .820 & .650 & .483 & .339 & .830 & .758 & .511 & .483 \\
kNN & .772 & .664 & .388 & .328 & .792 & .699 & .439 & .354 \\
Decision Tree & .820 & .619 & .500 & .338 & .823 & .763 & .516 & .527 \\
SVM (RBF) & .800 & .728 & .435 & .428 & .805 & .744 & .456 & .442 \\
MLP & .757 & .624 & .387 & .295 & .788 & .654 & .451 & .304 \\
Random Forest & .838 & .741 & .526 & .465 & .839 & .749 & .531 & .467 \\
XGBoost & .831 & .746 & .522 & .477 & .814 & .747 & .508 & .460 \\
\bottomrule
\end{tabular}
\end{table}

Fig.~\ref{fig:smote} shows the effect of SMOTE on binary MCC. It is negative for six of the seven models: largest for the MLP ($-0.064$) and kNN ($-0.051$), and negligible for Random Forest ($-0.004$). Only XGBoost gains ($+0.014$). On the tier scheme SMOTE costs Logistic Regression and the Decision Tree 14.4 and 18.9 MCC points.

\subsection{Feature ablation}

Table~\ref{tab:ablation} varies the feature set of a Random Forest held at the modal configuration of Table~\ref{tab:modal} (binary, with SMOTE). The configuration was chosen from the outer-fold searches, so these figures are mildly optimistic. Raw channels alone (A) score close to the full set (C). The temporal block alone (B) is clearly worse (MCC 0.366). However, B contains no $V_a$ term, because the rolling features are built from $P_g$, $I_a$, $V_g$ and $F_g$. Its deficit therefore reflects the absence of $V_a$ as much as the use of temporal features. AUC-ROC varies little (0.858--0.870), so the feature set mainly moves default-threshold decisions.

\begin{table}[t]
\centering
\caption{Random Forest feature ablation (binary, with SMOTE, fixed modal hyperparameters, 15 folds).}
\label{tab:ablation}
\small
\begin{tabular}{@{}lcccc@{}}
\toprule
Configuration & Features & Accuracy & MCC & AUC-ROC \\
\midrule
A: raw channels & 7 & .821$\pm$.171 & .504$\pm$.423 & .870$\pm$.201 \\
B: rolling and difference features (no $V_a$) & 18 & .743$\pm$.092 & .366$\pm$.295 & .860$\pm$.223 \\
C: all features & 29 & .839$\pm$.156 & .528$\pm$.417 & .858$\pm$.260 \\
\bottomrule
\end{tabular}
\end{table}

\subsection{Per-fault-type and per-episode performance}
\label{sec:perfault}

Table~\ref{tab:perfault} gives the binary Random Forest's recall per fault code (modal configuration, pooled out-of-fold predictions), together with the number of episodes detected. Because every code has between one and five episodes, the row-level recall is in effect an episode count, and each episode is classified almost all-or-nothing. Four of the 17 fault episodes are almost entirely missed (share of rows correct $\leq 0.007$): the single code-2.3 episode, one of the two uniform-shading episodes and two of the five intermittent episodes. One code-3.1 episode is partly missed (0.84). All other fault episodes are detected in at least 98\% of their rows.

\begin{table}[t]
\centering
\caption{Binary Random Forest per fault code (specificity for code 0.0), pooled over 15 folds. Episodes detected: share of rows correct $\geq 0.5$.}
\label{tab:perfault}
\small
\begin{tabular}{@{}llrcc@{}}
\toprule
Code & Description & Pooled rows & Recall & Episodes detected \\
\midrule
0.0 & No fault (specificity) & 42{,}000 & 0.935 & -- \\
1.0 & Uniform shading & 2{,}400 & 0.501 & 1 / 2 \\
2.1 & Constant partial, 1 module & 1{,}200 & 0.999 & 1 / 1 \\
2.2 & Constant partial, 2 modules & 1{,}200 & 1.000 & 1 / 1 \\
2.3 & Constant partial, 3 modules & 1{,}200 & 0.007 & 0 / 1 \\
3.1 & Constant partial, 1/3 module & 3{,}600 & 0.947 & 3 / 3 \\
3.2 & Constant partial, 2/3 module & 4{,}800 & 1.000 & 4 / 4 \\
4.0 & Intermittent (moving shadow) & 6{,}000 & 0.595 & 3 / 5 \\
\bottomrule
\end{tabular}
\end{table}

The missed episodes share an operating signature. The code-2.3 episode and one missed intermittent episode were recorded at low output (median $P_g$ of 217\,W and 179\,W), with $V_a$ of 181 and 169\,V. The no-fault median $V_a$ is 184\,V, against 151\,V for fault rows, so these episodes sit inside the no-fault voltage range. The other missed intermittent episode has near-normal $V_a$ (176\,V) at full output. A moving shadow may simply have been off the array for much of the retained 400-row slice, which is 2--15\% of each intermittent episode. Both uniform-shading episodes inject no power during their slices; the detected one also has $V_a=0$, whereas the missed one holds $V_a$ at about 150\,V, a value typical of fault rows at full output. Code 2.1, with fewer raw rows than code 2.3, is detected perfectly, which rules out sparsity as the explanation for code 2.3.

At tier level (three-class Random Forest with SMOTE, pooled; accuracy 0.768, macro-F1 0.564), the per-tier F1 is 0.909 for no fault, 0.587 for constant partial shading and 0.197 for severe shading (precision 0.272, recall 0.154). The faults most closely tied to the fire-risk motivation are the ones detected least reliably.

\section{Discussion}

\subsection{Comparative performance (RQ1)}

The ensembles have the best point estimates. The only differences that survive multiplicity correction are Random Forest's accuracy advantages over kNN and the MLP. Logistic Regression and the Decision Tree fall between the two groups and cannot be separated from either. A defensible reading is that the ensembles are at least as good as any alternative tested, and clearly better than kNN and the MLP. A clean split into two tiers is not supported.

\subsection{Effect of SMOTE (RQ2)}

SMOTE lowered binary MCC for six classifiers and helped only XGBoost, and then marginally. On the tier scheme it reversed the ranking of the Decision Tree and Logistic Regression relative to the ensembles. A likely reason is that fault episodes are long runs of near-identical readings, so interpolating between them adds little information about the class boundary while distorting distance-based, margin-based and single-tree learners. SMOTE should not be applied by default at the benchmark's 32.7\% prevalence, in line with reports from machine-fault and wind-turbine SCADA data \citep{Swana2022,MaldonadoCorrea2024}.

\subsection{Feature influence (RQ3)}

Once attributions are put on a common scale, $V_a$ is the dominant input for most models. A physical mechanism exists: a shaded cell limits the string current and triggers a bypass-diode voltage step, and the resulting multi-peak power--voltage curve can lead maximum power point tracking to a lower-voltage operating point \citep{Kappler2024}. The same direction of change, however, is produced by module heating at high irradiance, and in this benchmark the fault rows are concentrated at midday while the no-fault rows are not (Section~\ref{sec:confound}). The present data cannot separate the two mechanisms.

\subsection{Interpretability and operation (RQ4)}

The ensembles show no detectable gain over Logistic Regression or the Decision Tree. Both are cheaper to retrain and easier to audit: a depth-3 tree can be read as three threshold rules. For an operations team with limited data-science support, the evidence here does not justify the added opacity of an ensemble. That conclusion rests on non-significance in a small-sample design, not on demonstrated equivalence, and it should be revisited on matched data.

\subsection{Variability across folds}

Each outer fold holds out three or four fault episodes, and episodes are detected almost all-or-nothing. A fold's score therefore depends mainly on whether it contains one of the four hard episodes. This is why the standard deviations are large, and why paired comparison on identical folds is necessary.

\subsection{Operating-condition confound}
\label{sec:confound}

The subsampling keeps a contiguous 400-row slice (33--53\,min) per fault episode and a 2{,}000-row slice per no-fault run. All fault slices fall between 10:00 and 17:00. Most no-fault slices extend into early morning or late afternoon, between 05:51 and 18:11. The median clock hour is similar (12.6 versus 13.4), but the spread is not (s.d.\ 1.7 versus 3.7\,h), and the median grid power differs widely (982\,W for fault rows versus 274\,W for no-fault rows). To measure the effect, we re-ran the outer protocol with the modal Random Forest and no SMOTE on restricted inputs (Table~\ref{tab:confound}). Clock time alone, which carries no electrical information, gives MCC 0.44, more than kNN or the MLP achieve with 29 features. $V_a$ alone gives 0.72, and the six raw channels without $E_g$ give 0.74; both exceed every tuned model in Table~\ref{tab:primary}.

\begin{table}[t]
\centering
\caption{Post-hoc diagnostic: Random Forest (modal configuration, no SMOTE) on restricted inputs; same 3$\times$5 grouped outer folds.}
\label{tab:confound}
\small
\begin{tabular}{@{}lccc@{}}
\toprule
Inputs & Accuracy & MCC & AUC-ROC \\
\midrule
Clock hour only & .732$\pm$.114 & .443$\pm$.260 & .792 \\
$E_g$ only & .616$\pm$.130 & .196$\pm$.269 & .735 \\
$V_a$ only & .880$\pm$.075 & .724$\pm$.158 & .913 \\
$P_g$, $I_a$ & .822$\pm$.091 & .592$\pm$.189 & .821 \\
Seven raw channels & .851$\pm$.115 & .685$\pm$.217 & .876 \\
Six raw channels (no $E_g$) & .888$\pm$.084 & .744$\pm$.185 & .922 \\
\bottomrule
\end{tabular}
\end{table}

Two conclusions follow. First, part of what the classifiers learn is when and at what output level each class was recorded, not only the shading signature. The absolute scores in this paper should not be read as field detection rates. Second, the engineered features and SMOTE lowered the performance of the Random Forest instead of raising it. The ranking of the 29-feature models partly reflects how each model copes with redundant or confounded inputs. The remedy is to draw no-fault rows from the same clock window and irradiance band as the fault episodes, which the plant's companion meteorological file allows, and to replace $E_g$ by its rate of change.

\subsection{Validity and scope}

Three controls guard against train/test leakage: scaling and SMOTE are fitted on training folds only, tuning is nested, and no episode is split across folds. Every one of the 28 runs is reported. Pairwise tests are Holm-corrected, although they still treat correlated repeats as independent.

Three further limitations remain. The tier grouping is checked only against the binary scheme. The SHAP analysis explains default-parameter models in-sample. Tables~\ref{tab:ablation} and~\ref{tab:perfault} use a configuration chosen with knowledge of the outer folds.

Externally, the data come from one plant, one climate and one staging campaign, with all faults inside a 100-day window. The results are evidence about relative behaviour on this record under the stated sampling. They are not performance figures to expect at other installations.

\subsection{Deployment}

A deployed detector would inform maintenance dispatch and, for severe shading, a safety-relevant decision. The per-episode results show that a system built on these models would miss a materially larger share of uniform and intermittent shading than of constant partial shading. Before any operational use, the models would need validation on other plants and module technologies, a human-in-the-loop dispatch process, and monitoring for drift from soiling, degradation and season. None of these steps is undertaken here.

\section{Conclusions}

We compared seven classical and ensemble-tree classifiers on real, physically staged shading faults, under episode-grouped nested cross-validation with corrected paired tests. Random Forest and XGBoost gave the best point estimates. After multiplicity correction, only Random Forest's accuracy advantage over kNN and the MLP was significant, and Logistic Regression and the Decision Tree could not be distinguished from the ensembles. SMOTE reduced MCC for six of seven models and should not be a default. Errors concentrate in a few whole episodes, including half of the uniform-shading and two-fifths of the intermittent-shading episodes. A post-hoc diagnostic shows that the no-fault class differs from the fault class in time of day and output level, enough for clock time alone to reach an MCC of 0.44. The scores therefore mix shading signatures with acquisition context.

Future work will (i) re-run the benchmark with no-fault rows matched to the fault episodes in clock time and irradiance, using the companion meteorological data; (ii) replace the Wilcoxon tests with a corrected resampled $t$-test and report per-episode results; (iii) recompute SHAP on the tuned, out-of-fold models on a common output scale; (iv) extend the ablation to all feature blocks with and without SMOTE; and (v) evaluate sequence models and other real fault-labelled PV datasets as they become available.

\section*{Declaration of generative AI and AI-assisted technologies}
During the preparation of this work the authors used Claude (Anthropic) to assist with writing, drafting and copy-editing the manuscript, and to cross-check the reported figures against the stored results. All experiments were designed, and all scripts written and run, by the first author. The authors reviewed and edited all AI-assisted text against the experimental outputs and take full responsibility for the content of this publication.

\section*{CRediT authorship contribution statement}
\textbf{Zablon Ombiro:} Conceptualization, Methodology, Software, Validation, Formal analysis, Investigation, Data curation, Writing -- original draft, Writing -- review \& editing, Visualization. \textbf{Bernard Shibwabo:} Supervision, Validation, Writing -- review \& editing.

\section*{Declaration of competing interest}
The authors declare no known competing financial interests or personal relationships that could have influenced the work reported in this paper.

\section*{Data and code availability}
The dataset is publicly available on Zenodo \citep{Graillet2022}. The data-preparation, modelling and analysis code, the stored results and the post-hoc diagnostics (\texttt{review/reanalysis.py}) are available at \url{https://github.com/IncognitoGK9/pv-fault-benchmark} \citep{Ombiro2026}.

\section*{Acknowledgements}
The authors thank the Department of Computing, Strathmore University, for academic guidance and computing resources.

\section*{Funding}
This research received no external funding.

\end{document}